\documentclass[runningheads]{llncs}

\usepackage{graphicx}
\usepackage{amssymb}
\usepackage{amsmath}
\usepackage{mathtools}
\usepackage{tabularx}
\usepackage{booktabs}
\usepackage{longtable}
\usepackage{xcolor}
\usepackage{subcaption}
\usepackage[ruled,linesnumbered,commentsnumbered]{algorithm2e}
\usepackage{etoolbox}

\usepackage{lineno}

\newcolumntype{Y}{>{\centering\arraybackslash}X}

\DeclarePairedDelimiter\abs{\lvert}{\rvert}%
\DeclarePairedDelimiter\norm{\lVert}{\rVert}%
\makeatletter
\let\oldabs\abs
\def\abs{\@ifstar{\oldabs}{\oldabs*}}
\let\oldnorm\norm
\def\norm{\@ifstar{\oldnorm}{\oldnorm*}}
\makeatother

\begin{document}

\title{LipschitzLR: Using theoretically computed adaptive learning rates for fast convergence}
\titlerunning{LipschitzLR}

\author{Rahul Yedida \inst{1},  Snehanshu Saha \inst{2} \and Tejas Prashanth \inst{3}} 

\institute{Department of Computer Science, North Carolina State University, Raleigh, USA \email{ryedida@ncsu.edu} \and Department of CS\&IS and APPCAIR, BITS Pilani K K Birla Goa Campus, India \email{snehanshus@goa.bits-pilani.ac.in} \and Department of Computer Science, PES University, India\\ 
\email{tejuprash@gmail.com}}

\maketitle

\begin{abstract}
We present a novel theoretical framework for computing large, adaptive learning rates. Our framework makes minimal assumptions on the activations used and exploits the functional properties of the loss function. Specifically, we show that the inverse of the Lipschitz constant of the loss function is an ideal learning rate. We analytically compute formulas for the Lipschitz constant of several loss functions, and through extensive experimentation, demonstrate the strength of our approach using several architectures and datasets. In addition, we detail the computation of learning rates when other optimizers, namely, SGD with momentum, RMSprop, and Adam, are used. Compared to standard choices of learning rates, our approach converges faster, and yields better results.
\end{abstract}



\section{Introduction}
\label{sec:intro}
Gradient descent\cite{cauchy1847methode} is a popular optimization algorithm for finding optima for functions, and is used to find optima in loss functions in machine learning tasks. In an iterative process, it seeks to update randomly initialized weights to minimize the training error. These updates are typically small values proportional to the gradient of the loss function. The constant of proportionality is called the learning rate, and is usually manually chosen in the gradient descent rule.

When optimizing a function $f$ with respect to a parameter $\textbf{w}$, the gradient descent update rule is given by

\begin{equation}
    \textbf{w} := \textbf{w} - \alpha \cdot \nabla_{\textbf{w}} f
\end{equation}

The generalization ability of stochastic gradient descent (SGD) and various methods of faster optimization have quickly gained interest in machine learning and deep learning communities. 

Several directions have been taken to understand these phenomena. The interest in the stability of SGD is one such direction\cite{kuzborskij2017data} \cite{hardt2015train}. Others have proven that gradient descent can find the global minima of the loss functions in over-parameterized deep neural networks  \cite{zou2018stochastic} \cite{du2018gradient}.

More practical approaches in this regard have involved novel changes to the optimization procedure itself. These include adding a ``momentum" term to the update rule \cite{sutskever2013importance}, and ``adaptive gradient" methods such as RMSProp\cite{tieleman2012lecture}, and Adam\cite{kingma2014adam}. These methods have seen widespread use in deep neural networks\cite{radford2015unsupervised} \cite{xu2015show} \cite{bahar2017empirical}.  Other methods rely on an approximation of the Hessian. These include the Broyden-Fletcher-Goldfarb-Shanno (BFGS) \cite{broyden1970convergence} \cite{fletcher1970new} \cite{goldfarb1970family} \cite{shanno1970conditioning} and L-BFGS\cite{liu1989limited} algorithms. However, our proposed method does not require any modification of the standard gradient descent update rule, and only schedules the learning rate. Furthermore, for classical machine learning models, this learning rate is fixed and thus, our approach does not take any extra time. In addition, we only use the first gradient, thus requiring functions to be only once differentiable and $L$-Lipschitz.

Recently, several adaptive learning rate methods have been proposed. Zeiler \cite{zeiler2012adadelta} proposed AdaDelta, a per-dimension learning rate scheme that relies on computing the Hessian of the loss function. Zhou et al. \cite{zhou2018adashift} propose AdaShift, an improvement over the Adam algorithm. Wu et al. \cite{wu2018wngrad} propose WNGrad, an adaptive learning rate based on weight normalization. Reddi et al. \cite{reddi2019convergence} propose AMSGrad improve upon Adam by incorporating long-term memory into the update. Luo et al. \cite{luo2019adaptive} propose AdaBound and AMSBound, which rely on estimating the moment to find an adaptive learning rate.

However, we note several issues with the above approaches. For example, several methods rely on computing the Hessian of the loss function, which can be computationally expensive, and is typically estimated, using methods (for an example of such a method, see Martens et al. \cite{martens2012estimating}). In addition, a general trend that can be observed is to improve upon previous algorithms such as Adam. Finally, some of the methods above are specifically tuned for neural networks, and as such, are limited to deep learning.

Our approach alleviates the issues discussed above. Because our approach does not rely on computing the Hessian, it is computationally cheaper. Computing the first order derivatives (which is a significantly weaker condition to hold for commonly used loss functions) suffices for our approach. Further, as will be discussed in Section \ref{sec:practical}, modern frameworks include functions that allow us to compute our lipschitz adaptive learning rate in a significantly faster way. Our approach is also algorithm-independent; we describe our adaptive learning rate approach for SGD, SGD with momentum, and Adam. Further, while Wu et al. \cite{wu2018wngrad} note that "Thus in the stochastic setting, there is no 'best choice' for the learning rate", we disagree. Our learning rate is computed taking the mini-batch data into consideration, and therefore is truly adaptive in that it is tailored to each mini-batch. Finally, we argue in this paper that larger learning rates than typically used can work well, in agreement with Smith and Topin \cite{smith2017super}.

\subsection{Deep learning}
Deep learning \cite{goodfellow2016deep} is becoming more omnipresent for several tasks, including image recognition and classification \cite{simonyan2014very} \cite{szegedy2015going} \cite{sermanet2013overfeat} \cite{zeiler2014visualizing}, face recognition \cite{taigman2014deepface}, and object detection \cite{girshick2014rich}, even surpassing human-level performance\cite{he2015delving}.  At the same time, the trend is towards deeper neural networks \cite{ioffe2015batch} \cite{he2015delving}.

Despite their popularity, training neural networks is made difficult by several problems. These include vanishing and exploding gradients \cite{glorot2010understanding} \cite{bengio1994learning} and overfitting. Various advances including different activation functions \cite{klambauer2017self} \cite{nair2010rectified}, batch normalization \cite{ioffe2015batch}, novel initialization schemes \cite{he2015delving}, and dropout \cite{srivastava2014dropout} offer solutions to these problems. 

However, a more fundamental problem is that of finding optimal values for various hyperparameters, of which the learning rate is arguably the most important. It is well-known that learning rates that are too small are slow to converge, while learning rates that are too large cause divergence \cite{bengio2012neural}. Recent works agree that rather than a fixed learning rate value, a non-monotonic learning rate scheduling system offers faster convergence \cite{seong2018towards} \cite{smith2017cyclical}. It has also been argued that the traditional wisdom that large learning rates should not be used may be flawed, and can lead to ``super-convergence" and have regularizing effects \cite{smith2017super}. Our experimental results agree with this statement; however, rather than use cyclical learning rates based on intuition, we propose a novel method to compute an adaptive learning rate backed by theoretical foundations.

Recently, there has been a lot of work on finding novel ways to adaptively change the learning rate. These have both theoretical \cite{seong2018towards} and intuitive, empirical \cite{smith2017super} \cite{smith2017cyclical} backing. These works rely on non-monotonic scheduling of the learning rate. \cite{smith2017cyclical} argues for cyclical learning rates. Our proposed method also yields a non-monotonic learning rate, but does not follow any predefined shape.
\section{Our Contribution}
In this paper, we propose a novel theoretical framework to compute large, adaptive learning rates for use in gradient-based optimization algorithms. We start with a presentation of the theoretical framework and the motivation behind it, and then derive the mathematical formulas to compute the learning rate on each epoch. We then extend our approach from stochastic gradient descent (SGD) to other optimization algorithms. Finally, we present extensive experimental results to support our claims. 

Our experimental results show that compared to standard choices of learning rates, our approach converges quicker and achieves better results. During the experiments, we explore cases where adaptive learning rates outperform fixed learning rates. Our approach exploits functional properties of the loss function, and only makes two minimal assumptions on the loss function: it must be Lipschitz continuous\cite{saha} and (at least) once differentiable. Commonly used loss functions satisfy both these properties. 

In summary, our contributions in this paper are as follows:
\begin{itemize}
    \item We present a theoretical framework based on the Lipschitz constant of the loss function to compute an adaptive learning rate.
    \item We provide an intuitive motivation and mathematical justification for using the inverse of the Lipschitz constant as the learning rate in gradient-based optimization algorithms, and derive formulas for the Lipschitz constant of several commonly used loss functions. We note that classical machine learning models, such as logistic regression, are simply special cases of deep learning models, and show the equivalence of the formulas derived.
    \item We argue that the use of Lipschitz constants to determine learning rate accelerates convergence many-fold in comparison with standard learning rate choices. We present empirical evidence of our claims in Section 6. This is a departure from the approach of manually tuning learning rates.
    \item Through extensive experimentation, we demonstrate the strength of our results with both classical machine learning models and deep learning models.
\end{itemize}
The rest of the paper is organized as follows. The paper begins with our theoretical framework based on the properties of several loss functions, in Section 3. Section 4 provides mathematical justification for setting the learning rate as the inverse of the Lipschitz constant and argues that such setting shall provide an estimate for number of iterations to converge to the global optima. Section 5 derives the Lipschitz constant for regression problems. Section 6 deals with derivation of learning rate in classification problems. In Section 7, we extend the framework to algorithms that extend SGD, such as RMSprop, momentum, and Adam. Section 8 details our experiments and discusses their results. Section 9 discusses some practical considerations of our approach. We conclude in section 10 with a brief discussion.\\

\section{Theoretical Framework}
\label{sec:theory}

\subsection{Introduction and Motivation}
For a function, the Lipschitz constant is the least positive constant $L$ such that 

\begin{equation}
    \left\Vert f(\textbf{w}_1) - f(\textbf{w}_2)\right\Vert \leq L \left\Vert \textbf{w}_1 - \textbf{w}_2 \right\Vert
\end{equation}

for all $\textbf{w}_1$, $\textbf{w}_2$ in the domain of $f$. From the mean-value theorem for scalar fields, for any $\textbf{w}_1, \textbf{w}_2$, there exists $\textbf{v}$ such that 

\[
    \begin{aligned}
        \norm{f(\textbf{w}_1) - f(\textbf{w}_2)} &= \norm{ \nabla_{\textbf{w}} f(\textbf{v})} \norm{ \textbf{w}_1-\textbf{w}_2} \\
        &\leq \sup\limits_{\textbf{v}} \norm{\nabla_{\textbf{w}} f(\textbf{v})} \norm{\textbf{w}_1-\textbf{w}_2}
    \end{aligned}
\]
Thus, $\sup\limits_{\textbf{v}} \norm{\nabla_{\textbf{w}} f(\textbf{v})}$ is such an $L$. Since $L$ is the least such constant, 

\begin{equation}
    L \leq \sup\limits_{\textbf{v}} \norm{ \nabla_{\textbf{w}} f(\textbf{v}) }
\end{equation}

In this paper, we use $\max \norm{\nabla_{\textbf{w}} f}$ to derive the Lipschitz constants. Our approach makes the minimal assumption that the functions are Lipschitz continuous and differentiable up to first order only \footnote{Note this is a weaker condition than assuming the gradient of the function being Lipschitz continuous. We exploit merely the boundedness of the gradient.}. Because the gradient of these loss functions is used in gradient descent, these conditions are guaranteed to be satisfied. 

By setting $\alpha = \frac{1}{L}$, we have $\Delta \textbf{w} \leq 1$, constraining the change in the weights. We stress here that we are not computing the Lipschitz constants of the \textit{gradients} of the loss functions, but of the losses themselves. Therefore, our approach merely assumes the loss is $L$-Lipschitz, and not $\beta$-smooth. We argue that the boundedness of the effective weight changes makes it optimal to set the learning rate to the reciprocal of the Lipschitz constant. This claim, while rather bold, is supported by our experimental results.
\section{Mathematical Justification of the Learning Rate Setting}
A typical Gradient Descent algorithm is in the form of 
\begin{align}
    \boldsymbol{w^{k+1}}= \boldsymbol{w^{k}} - \eta_{k} \nabla_{w} f \label{weight_update_rule}
\end{align} 
   for $k \in \mathbb{Z}$
    and stop if $\left\Vert \nabla_{w} f \right\Vert \leq \epsilon$
    
$\forall v,w, \exists L$ such that $\left\Vert f(\textbf{v}) - f(\textbf{w})\right\Vert \leq L \left\Vert \textbf{v} - \textbf{w} \right\Vert$. The assumption that gradients cannot change arbitrarily fast is fairly weak. So, for $f \in C^{2}$, the Lipschitz condition can be written as  $\nabla^{2} f(w) \leq LI$, where $I$ represents the Identity matrix and $L$ be the Lipschitz constant (Note, for scalar functions, $f \in C^{2},  \nabla^{2} f(w) \leq L$). $\nabla^{2} f(w) \leq LI \implies$ 
\begin{align}
    \boldsymbol{v^{T}} \boldsymbol{\nabla^{2}} \boldsymbol{f(u)} \boldsymbol{v} \leq \boldsymbol{v^{T}} (L\boldsymbol{I}) \boldsymbol{v} => 
    \boldsymbol{v^{T}} \nabla^{2} \boldsymbol{f(u)} \boldsymbol{v} \leq L \left\Vert \boldsymbol{v} \right\Vert^{2} \label{second_derivative_condition}
\end{align}{}
for any $u,v, w$ where $u$ is a convex combination of $v, w$. By the Taylor series and using \eqref{second_derivative_condition},
\begin{align}
    \boldsymbol{f(v)} = \boldsymbol{f(w)} + \nabla\boldsymbol{f(w)^{T}} (\boldsymbol{v-w}) + \frac{1}{2} (\boldsymbol{v-w})^{T} \nabla^{2} \boldsymbol{f(w)} (\boldsymbol{v-w}) \nonumber \\
    \boldsymbol{f(v)} \leq \boldsymbol{f(w)} + \nabla\boldsymbol{f(w)^{T}} (\boldsymbol{v-w}) + \frac{L}{2} \left\Vert \boldsymbol{v-w} \right\Vert^{2} \label{taylor_series_inequality}
\end{align}{}
This provides a convex quadratic upper bound which can be minimized using gradient descent with the learning rate $\eta_{k}$ = $\frac{1}{L}$.
Consider 
$\boldsymbol{w^{k+1}}= \boldsymbol{w^{k}} - \eta_{k} \nabla_{w} f $.  
Substituting $\boldsymbol{w^{k}}$ and $\boldsymbol{w^{k+1}}$ for $\boldsymbol{w}$ and $\boldsymbol{v}$ respectively in \ref{taylor_series_inequality} and using \ref{weight_update_rule}, we obtain

\begin{align*}
    \boldsymbol{f(w^{k+1})} \leq \boldsymbol{f(w^{k})} + \nabla\boldsymbol{f(w^{k})^{T}} (\boldsymbol{w^{k+1}-w^{k}}) + \frac{L}{2} \left\Vert \boldsymbol{w^{k+1}-w^{k}} \right\Vert^{2} \\
    \boldsymbol{f(w^{k+1})} \leq \boldsymbol{f(w^{k})} - \frac{1}{L} \nabla\boldsymbol{f(w^{k})^{T}} \nabla\boldsymbol{f(w^{k})} + \frac{L}{2} \left\Vert  \frac{-1}{L} \boldsymbol{\nabla f(w^{k}) } \right\Vert^{2} \\
    \boldsymbol{f(w^{k+1})} \leq \boldsymbol{f(w^{k})} - \frac{1}{L} \left\Vert \nabla\boldsymbol{f(w^{k})} \right\Vert^{2} + \frac{1}{2L} \left\Vert  \boldsymbol{\nabla f(w^{k}) } \right\Vert^{2} \\
    \boldsymbol{f(w^{k+1})} \leq \boldsymbol{f(w^{k})} - \frac{1}{2L} \left\Vert \nabla\boldsymbol{f(w^{k})} \right\Vert^{2} 
\end{align*}{}

Therefore, Gradient Descent decreases $\boldsymbol{f(w)}$ if $\eta_{k}=\frac{1}{L}$ and $\eta_{k} < \frac{2}{L}$. This proof also enables one to derive the rate of convergence with Lipschitz adaptive learning rate. 

\subsection{Convergence rate}
A choice of learning rate $\eta_{k}=\frac{1}{L}$ and $\eta_{k} < \frac{2}{L}$ ensures that $f(w)$ converges to a global optimum. In order to prove that, we must prove that $\left\Vert \nabla f(w^{k}) \right\Vert \leq \epsilon$ such that the rate of convergence can be found. The following assumptions are made, which are consistent with the computation of Lipschitz constant- $\nabla f$ is Lipschitz continuous, the gradient descent rule follows a step size,$\eta_{k} = \frac{1}{L} $ and $f$ is bounded below. Most loss functions satisfy all of these properties.

Let $n$ represent the number of iterations. We know that,
\begin{align*}
        \boldsymbol{f(w^{k+1})} \leq \boldsymbol{f(w^{k})} - \frac{1}{2L} \left\Vert \nabla\boldsymbol{f(w^{k})} \right\Vert^{2} \\
        \left\Vert \nabla\boldsymbol{f(w^{k})} \right\Vert^{2} \leq 2L (\boldsymbol{f(w^{k})} - \boldsymbol{f(w^{k+1})}) \\
        \sum_{k=1}^{n} \left\Vert \nabla\boldsymbol{f(w^{k})} \right\Vert^{2} \leq 2L \sum_{k=1}^{n} (\boldsymbol{f(w^{k})} - \boldsymbol{f(w^{k+1})}) \\
        \sum_{k=1}^{n} \left\Vert \nabla\boldsymbol{f(w^{k})} \right\Vert^{2} \leq 2L (\boldsymbol{f(w^{0})} - \boldsymbol{f(w^{n+1})})
\end{align*}{}

The RHS of the above inequality follows from a simple telescoping sum. Moreover, since
\begin{align*}
    \sum_{k=1}^{n} \min_{j \in {1,2,..n}} {\left\Vert \nabla\boldsymbol{f(w^{j})} \right\Vert^{2}} \leq \sum_{k=1}^{n} \left\Vert \nabla\boldsymbol{f(w^{k})} \right\Vert^{2}
\end{align*}{}

It follows that,
\begin{align*}
    \sum_{k=1}^{n} \min_{j \in {1,2,..n}} {\left\Vert \nabla\boldsymbol{f(w^{j})} \right\Vert^{2}} \leq 2L (\boldsymbol{f(w^{0})} - \boldsymbol{f(w^{n+1})}) \\
    n \min_{j \in {1,2,..n}} {\left\Vert \nabla\boldsymbol{f(w^{j})} \right\Vert^{2}} \leq 2L (\boldsymbol{f(w^{0})} - \boldsymbol{f(w^{n+1})}) \\
    n \min_{j \in {1,2,..n}} {\left\Vert \nabla\boldsymbol{f(w^{j})} \right\Vert^{2}} \leq 2L (\boldsymbol{f(w^{0})} - \boldsymbol{f(w^{*})})   \text{where} f(w^{*}) < f(w^{n+1}) \\
    \min_{j \in {1,2,..n}} {\left\Vert \nabla\boldsymbol{f(w^{j})} \right\Vert^{2}} \leq \frac{2L (\boldsymbol{f(w^{0})} - \boldsymbol{f(w^{*})})}{n} = O(\frac{1}{n})\\
\end{align*}{}
If,
\begin{align*}
    \frac{2L (\boldsymbol{f(w^{0})} - \boldsymbol{f(w^{*})})}{n} \leq \epsilon \\
    n \geq  \frac{2L (\boldsymbol{f(w^{0})} - \boldsymbol{f(w^{*})})}{\epsilon}
\end{align*}{}
Therefore, Gradient Descent requires, at least, $n = O(\frac{1}{\epsilon})$ iterations to achieve  the error bound:
$\left\Vert \nabla f(w^{k}) \right\Vert \leq \epsilon$ \footnote{Since $f \in C^2$ is a condition for obtaining lower bound on the number of iterations to converge for the choice of Lipschitz learning rate, Mean Absolute Error can't be used as loss function.}
\subsection{Significance of Lipschitz constant (LC)}
The Lipschitz constant (LC) has found a variety of uses in computing and applications. The central condition to the existence and uniqueness of solutions to first order system of differential equations of the form $y'(t)=f(t,y(t)$ is LC of $f$. The existence of LC guarantees contraction and eventually a fixed point i.e. solution to the above system\cite{banach1922operations} and saves the trouble of computing an analytical solution to the system above. Finding an LC is equivalent to to the fact that the function, $f$ possesses Lipschitz continuity. Given, $f:R\xrightarrow{}R, $ there exists an $L$ such that
\begin{equation*}
    \left\Vert f(\textbf{x}) - f(\textbf{y})\right\Vert \leq L \left\Vert \textbf{x} - \textbf{y} \right\Vert
\end{equation*}
Consequently, \begin{equation*}
   \frac{\left\Vert f(\textbf{x}) - f(\textbf{y})\right\Vert}{\left\Vert \textbf{x} - \textbf{y} \right\Vert}  \leq L \end{equation*} 
This implies, the slope of the secant line connecting $x,y$ is bounded above. This is equivalent to stating that computing a LC of a function (loss function, in our case) is identical to computing the maximum of the derivative of $f$. This also establishes the relation between finding LC and the Mean Value Theorem. Note, however, that Lipschitz continuity is not synonymous to uniform continuity, a much stronger condition. For loss functions which are differentiable, we can easily compute LCs and therefore find the bound on the derivatives to be used in deep neural network training. Our paper focuses on loss functions that satisfy the conditions of differentiability and hence Lipschitz continuity. We compute LCs of those loss functions in the subsequent sections and arrive at adaptive learning rate formulation.
\subsection{Notation}
We use the following notation:
\begin{itemize}
    \item $(x^{(i)}, y^{(i)})$ refers to one training example. The superscript with parentheses indicates the $i$th training example. $X$ refers to the input matrix.
    \item Where not specified, it should be assumed that $m$ indicates the number of training examples.
    \item For deep neural networks, whenever unclear, we use a superscript with square brackets to indicate the layer number. For example, $W^{[l]}$ indicates the weight matrix at the $l$th layer. We use $L$ to represent the total number of layers, being careful not to cause ambiguity with the Lipschitz constant.
    \item We use the letter $w$ or $W$ to refer to weights, while $b$ refers to a bias term. Capital letters indicate matrices; lowercase letters indicate scalars, and are usually accompanied by subscripts--in such a case, we will adequately describe what the subscripts indicate.
    \item We use the letter $a$ to denote an activation; thus, $a^{[l]}$ represents the activations at the $l$th layer.
    \item Where not specified, the matrix norm is the Frobenius norm, and $\norm{z} = z$ when $z \in \mathbb{R}$.
\end{itemize}

\subsection{Deriving the Lipschitz constant for neural networks}

For a neural network that uses the sigmoid, ReLU, or softmax activations, it is easily shown that the gradients get smaller towards the earlier layers in backpropagation. Because of this, the gradients at the last layer are the maximum among all the gradients computed during backpropagation. If $w^{[l]}_{ij}$ is the weight from node $i$ to node $j$ at layer $l$, and if $L$ is the number of layers, then

\begin{equation}
	\max\limits_{h, k} \norm{\frac{\partial E}{\partial w^{[L]}_{hk}}} \geq \norm{\frac{\partial E}{\partial w^{[l]}_{ij}}} \forall\  l, i, j \label{eq:int:1}
\end{equation}

Essentially, \eqref{eq:int:1} says that the maximum gradient of the error with respect to the weights in the last layer is greater than the gradient of the error with respect to any weight in the network. In other words, finding the maximum gradient at the last layer gives us a supremum of the Lipschitz constants of the error, where the gradient is taken with respect to the weights at any layer. We call this supremum as a Lipschitz constant of the loss function for brevity.

We now analytically arrive at a theoretical Lipschitz constant for different types of problems. The inverse of these values can be used as a learning rate in gradient descent. Specifically, since the Lipschitz constant that we derive is an upper bound on the gradients, we effectively limit the size of the parameter updates, without necessitating an overly guarded learning rate. In any layer, we have the computations
\begin{align}
	z^{[l]} &= W^{[l]T}a^{[l-1]} + b^{[l]} \label{eq:int:2} \\
	a^{[l]} &= g(z^{[l]}) \label{eq:int:3} \\
	a^{[0]} &= X \label{eq:int:4}
\end{align}
Thus, the gradient with respect to any weight in the last layer is computed via the chain rule as follows.
\begin{align}
	\frac{\partial E}{\partial w^{[L]}_{ij}} &= \frac{\partial E}{\partial a^{[L]}_j}\cdot \frac{\partial a^{[L]}_j}{\partial z^{[L]}_j}\cdot \frac{\partial z^{[L]}_j}{\partial w^{[L]}_{ij}} \nonumber  \\
	&= \frac{\partial E}{\partial a^{[L]}_j}\cdot \frac{\partial a^{[L]}_j}{\partial z^{[L]}_j}\cdot a^{[L-1]}_i \label{eq:int:5}
\end{align}
This gives us
\begin{equation}
	\max\limits_{i,j} \abs{\frac{\partial E}{\partial w^{[L]}_{ij}}} \leq \max\limits_j \abs{ \frac{\partial E}{\partial a^{[L]}_j} }\cdot \max\limits_j \abs{ \frac{\partial a^{[L]}_j}{\partial z^{[L]}_j} }\cdot \max\limits_j \abs{ a^{[L-1]}_j } \label{eq:int:6}
\end{equation}
The third part cannot be analytically computed; we denote it as $K_z$. We now look at various types of problems and compute these components. Note that we use the terms ``cost function" and ``loss function" interchangeably.

\section{Least-squares cost function}
\label{sec:leastsq}
For the least squares cost function, we will separately compute the Lipschitz constant for a linear regression model and for neural networks where the output is continuous. We will then prove the equivalence of the two results, deriving the former as a special case of the latter.

\subsection{Linear regression}
We have,
\[
    g(\textbf{w}) = \frac{1}{2m}\sum\limits_{i=1}^m \left(\textbf{x}^{(i)} \textbf{w} - y^{(i)}\right)^2
\]
Thus,
\[
    \begin{aligned}
        g(\textbf{w}) - g(\textbf{v}) &= \frac{1}{2m}\sum\limits_{i=1}^m \left(\textbf{x}^{(i)} \textbf{w} - y^{(i)}\right)^2 - \left(\textbf{x}^{(i)} \textbf{v} - y^{(i)}\right)^2 \\
        &= \frac{1}{2m}\sum\limits_{i=1}^m \left( \textbf{x}^{(i)}(\textbf{w}+\textbf{v}) - 2y^{(i)}\right) \left( \textbf{x}^{(i)} (\textbf{w}-\textbf{v}) \right) \\
        &= \frac{1}{2m}\sum\limits_{i=1}^m \left( (\textbf{w}+\textbf{v})^T \textbf{x}^{(i)T} - 2y^{(i)}\right) \left( \textbf{x}^{(i)} (\textbf{w}-\textbf{v}) \right) \\
        &= \frac{1}{2m}\sum\limits_{i=1}^m \left( (\textbf{w} + \textbf{v})^T \textbf{x}^{(i)T}\textbf{x}^{(i)} - 2y^{(i)}\textbf{x}^{(i)} \right) (\textbf{w}-\textbf{v}) 
    \end{aligned}
\]
The penultimate step is obtained by observing that $(\textbf{w}+\textbf{v})^T \textbf{x}^{(i)T}$ is a real number, whose transpose is itself.

At this point, we take the norm on both sides, and then assume that $\textbf{w}$ and $\textbf{v}$ are bounded such that $\left\Vert \textbf{w} \right\Vert, \left\Vert \textbf{v} \right\Vert \leq K$. Taking norm on both sides,
\[
    \boxed{
        \frac{\left\Vert g(\textbf{w}) - g(\textbf{v}) \right\Vert}{\left\Vert \textbf{w} - \textbf{v} \right\Vert} \leq \frac{K}{m}\left\Vert \textbf{X}^T\textbf{X} \right\Vert + \frac{1}{m} \left\Vert\textbf{y}^T \textbf{X} \right\Vert
    }
\]
We are forced to use separate norms because the matrix subtraction $2K \textbf{X}^T\textbf{X} - 2\textbf{y}^T \textbf{X}$ cannot be performed. The RHS here is the Lipschitz constant. Note that the Lipschitz constant changes if the cost function is considered with a factor other than $\frac{1}{2m}$.

\subsection{Regression with neural networks}
Let the loss be given by
\begin{equation}
	E(\textbf{a}^{[L]}) = \frac{1}{2m} \left( \textbf{a}^{[L]} - \textbf{y} \right)^2 \label{eq:reg:1}
\end{equation}
where the vectors contain the values for each training example. Then we have,
\begin{align*}
	E(\textbf{b}^{[L]}) - E(\textbf{a}^{[L]}) &= \frac{1}{2m} \left( \left( \textbf{b}^{[L]} - \textbf{y} \right)^2 - \left( \textbf{a}^{[L]} - \textbf{y} \right)^2 \right) \\
	&= \frac{1}{2m} \left( \textbf{b}^{[L]} + \textbf{a}^{[L]} - 2\textbf{y} \right) \left( \textbf{b}^{[L]} - \textbf{a}^{[L]} \right)
\end{align*}
This gives us,
\begin{align}
	\frac{\lVert E(\textbf{b}^{[L]}) - E(\textbf{a}^{[L]}) \rVert}{\lVert \textbf{b}^{[L]} - \textbf{a}^{[L]} \rVert} &= \frac{1}{2m} \lVert \textbf{b}^{[L]} + \textbf{a}^{[L]} - 2\textbf{y} \rVert \nonumber \\
	& \leq \frac{1}{m} \left( K_a + \norm{\textbf{y}} \right) \label{eq:reg:2}
\end{align}
where $K_a$ is the upper bound of $\norm{\textbf{a}}$ and $\norm{\textbf{b}}$. A reasonable choice of norm is the 2-norm.

Looking back at \eqref{eq:int:6}, the second term on the right side of the equation is the derivative of the activation with respect to its parameter. Notice that if the activation is sigmoid or softmax, then it is necessarily less than 1; if it is ReLu, it is either 0 or 1. Therefore, to find the maximum, we assume that the network is comprised solely of ReLu activations, and the maximum of this is 1.

From \eqref{eq:int:6}, we have
\begin{equation}
\boxed{
	\max\limits_{i,j} \norm{ \frac{\partial E}{\partial w^{[L]}_{ij}} } = \frac{1}{m} \left( K_a + \norm{\textbf{y}} \right) K_z
}
\label{eq:reg:nn}
\end{equation}

\subsection{Equivalence of the constants}
The equivalence of the above two formulas is easy to see by understanding the terms of \eqref{eq:reg:nn}. We had defined in \eqref{eq:int:6}, 
\begin{equation}
    K_z = \max\limits_j \norm{a_j^{[L-1]}}
\end{equation}
Because a linear regression model can be thought of as a neural network with no hidden layers and a linear activation, and from \eqref{eq:int:4}, we have, 
\[
    \textbf{a}^{[L-1]} = \textbf{a}^0 = \textbf{X}
\]
and therefore 
\begin{equation}
    K_z = \max\limits_j \norm{a_j^{[L-1]}} = \norm{\textbf{X}}
    \label{eq:reg:Kz}
\end{equation}
Next, observe that $K_a$ is the upper bound of the final layer activations. For a linear regression model, we have the ``activations" as the outputs: $\hat{\textbf{y}} = \textbf{W}^T \textbf{X}$. Using the assumption that $\norm{\textbf{W}}$ has an upper bound $K$, we obtain
\begin{equation}
    K_a = \max\limits \norm{\textbf{a}^{[L]}} = \max \norm{\textbf{W}^T \textbf{X}} = \max \norm{\textbf{W}} \cdot \norm{\textbf{X}} = K\norm{\textbf{X}}
    \label{eq:reg:Ka}
\end{equation}
Substituting \eqref{eq:reg:Kz} and \eqref{eq:reg:Ka} in \eqref{eq:reg:nn}, we obtain
\begin{align*}
    \max\limits_{i,j} \norm{ \frac{\partial E}{\partial w^{[L]}_{ij}} } &= \frac{1}{m} \left( K_a + \norm{\textbf{y}} \right) K_z \\
    &= \frac{1}{m}\left( K \norm{\textbf{X}} + \norm{\textbf{y}} \right) \norm{\textbf{X}} \\
    &= \frac{K}{m}\norm{\textbf{X}^T\textbf{X}} + \frac{1}{m}\norm{\textbf{y}^T \textbf{X}}
\end{align*}
$\hfill\square$

This argument can also be used for the other loss functions that we discuss below; therefore, we will not prove equivalence of the Lipschitz constants for classical machine learning models (logistic regression and softmax regression) and neural networks. However, we will show experiments on both separately.

\section{Classification}
\label{sec:classification}

\subsection{Binary classification}
For binary classification, we use the binary cross-entropy loss function. Assuming only one output node,
\begin{equation}
    E(\textbf{z}^{[L]}) = -\frac{1}{m} \left( \textbf{y} \log g(\textbf{z}^{[L]}) + (1-\textbf{y}) \log (1 - g(\textbf{z}^{[L]})) \right) \label{eq:bin:1}
\end{equation}
where $g(z)$ is the sigmoid function. We use a slightly different version of \eqref{eq:int:6} here:
\begin{equation}
    \max\limits_{i,j} \abs{ \frac{\partial E}{\partial w^{[L]}_{ij}} } = \max\limits_j \abs{ \frac{\partial E}{\partial z^{[L]}_j} }\cdot K_z \label{eq:bin:2}
\end{equation}
Then, we have
\begin{align}
    \frac{\partial E}{\partial \textbf{z}^{[L]}} &= -\frac{1}{m} \left( \frac{\textbf{y}}{g(\textbf{z}^{[L]})}g(\textbf{z}^{[L]})(1-g(\textbf{z}^{[L]})) - \frac{1-\textbf{y}}{1-g(\textbf{z}^{[L]})}g(\textbf{z}^{[L]})(1-g(\textbf{z}^{[L]})) \right) \nonumber \\
    &= -\frac{1}{m}\left( \textbf{y}(1-g(\textbf{z}^{[L]})) - (1-\textbf{y})g(\textbf{z}^{[L]}) \right) \nonumber \\
    &= -\frac{1}{m}\left( \textbf{y} - \textbf{y}g(\textbf{z}^{[L]}) - g(\textbf{z}^{[L]}) + \textbf{y}g(\textbf{z}^{[L]}) \right) \nonumber \\
    &= -\frac{1}{m}\left( \textbf{y} - g(\textbf{z}^{[L]}) \right) \label{eq:bin:3}
\end{align}
It is easy to show, using the second derivative, that this attains a maxima at $\textbf{z}^{[L]}=0$:
\begin{equation}
    \frac{\partial^2 E}{\partial \textbf{w}^{[L]2}_{ij}} = \frac{1}{m}g(\textbf{z}^{[L]})(1 - g(\textbf{z}^{[L]})) a^{[L-1]}_j \label{eq:bin:4}
\end{equation}
Setting \eqref{eq:bin:4} to 0 yields $a^{[L-1]}_j = 0\ \forall j$, and thus $z^{[L]} = W^{[L]}_{ij}a^{[L-1]}_j = 0$. This implies $g(\textbf{z}^{[L]}) = \frac{1}{2}$. Now whether $\textbf{y}$ is 0 or 1, substituting this back in \eqref{eq:bin:3}, we get
\begin{equation}
    \max_j \norm{ \frac{\partial E}{\partial z^{[L]}_j} } = \frac{1}{2m} \label{eq:bin:5}
\end{equation}
Using \eqref{eq:bin:5} in \eqref{eq:bin:2},
\begin{equation}
\boxed{
    \max\limits_{i,j} \norm{ \frac{\partial E}{\partial w^{[L]}_{ij}} } = \frac{K_z}{2m}
}
\label{eq:bin:6}
\end{equation}

We simply mention here that for logistic regression, the Lipschitz constant is
\[
    \boxed{
        L = \frac{1}{2m} \norm{\textbf{X}}
    }
\]

\subsection{Multi-class classification}
While conventionally, multi-class classification is done using one-hot encoded outputs, that is not convenient to work with mathematically. An identical form of this is to assume the output follows a Multinomial distribution, and then updating the loss function accordingly. This is because the effect of the typical loss function used is to only consider the ``hot" vector; we achieve the same effect using the Iverson notation, which is equivalent to the Kronecker delta. With this framework, the loss function is
\begin{equation}
     E(\textbf{a}^{[L]}) = -\frac{1}{m} \sum\limits_{j=1}^k [\textbf{y}=j] \log \textbf{a}^{[L]} \label{eq:mul:1}
\end{equation}
Then the first part of \eqref{eq:int:6} is trivial to compute:
\begin{equation}
    \frac{\partial E}{\partial \textbf{a}^{[L]}} = -\frac{1}{m} \sum\limits_{j=1}^m \frac{[\textbf{y}=j]}{\textbf{a}^{[L]}} \label{eq:mul:2}
\end{equation}
The second part is computed as follows.
\begin{align} 
    \frac{\partial a^{[L]}_j}{\partial z^{[L]}_p} &= \frac{\partial}{\partial z^{[L]}_p} \left( \frac{e^{z^{[L]}_j}}{\sum_{l=1}^k e^{z^{[L]}_l}} \right) \nonumber \\ 
    &= \frac{[p=j] e^{z^{[L]}_j}\sum_{l=1}^k e^{z^{[L]}_l } - e^{z^{[L]}_j } \cdot e^{z^{[L]}_p} }{\left( \sum_{l=1}^k e^{z^{[L]}_l } \right)^2} \nonumber \\ 
    &= \frac{[p=j] e^{z^{[L]}_j }}{\sum_{l=1}^k e^{z^{[L]}_l }} - \frac{e^{z^{[L]}_j }}{\sum_{l=1}^k e^{z^{[L]}_l }} \cdot \frac{e^{z^{[L]}_p }}{\sum_{l=1}^k e^{z^{[L]}_l }} \nonumber \\ 
    &= \left([p=j] a^{[L]}_j - a^{[L]}_j a^{[L]}_p \right) \nonumber \\ 
    &= a^{[L]}_j([p=j]-a^{[L]}_p) \label{eq:mul:3}
\end{align}
Combining \eqref{eq:mul:2} and \eqref{eq:mul:3} in \eqref{eq:int:5} gives
\begin{equation}
    \frac{\partial E}{\partial W^{[L]}_p} = \frac{1}{m} \left( a^{[L]}_p - [\textbf{y}=p] \right)K_z \label{eq:mul:4}
\end{equation}
It is easy to show that the limiting case of this is when all softmax values are equal and each $y^{(i)}=p$; using this and $a^{[L]}_p = \frac{1}{k}$ in \eqref{eq:mul:4} and combining with \eqref{eq:int:6} gives us our desired result:
\begin{equation}
\boxed{
    \max\limits_j \norm{ \frac{\partial E}{\partial W^{[L]}_j} } = \frac{k-1}{km}K_z
}
\label{eq:mul:5}
\end{equation}
For a softmax regression model, we have
\[
    \boxed{
        L = \frac{k-1}{km}\norm{\textbf{X}}
    }
\]

\subsection{Regularization}
\label{sec:regularization}

This framework is extensible to the case where the loss function includes a regularization term. 

In particular, if an $L_2$ regularization term, $\frac{\lambda}{2}\left\Vert \textbf{w} \right\Vert_2^2$ is added, it is trivial to show that the Lipschitz constant increases by $\lambda K$, where $K$ is the upper bound for $\left\Vert \textbf{w} \right\Vert$. More generally, if a Tikhonov regularization term, $\left\Vert \boldsymbol\Gamma \textbf{w} \right\Vert_2^2$ term is added, then the increase in the Lipschitz constant can be computed as below.

\[
    \begin{aligned}
        L(\textbf{w}_1) - L(\textbf{w}_2) &= (\boldsymbol\Gamma \textbf{w}_1)^T (\boldsymbol\Gamma \textbf{w}_1) - (\boldsymbol\Gamma \textbf{w}_2)^T (\boldsymbol\Gamma \textbf{w}_2) \\
        &= \textbf{w}_1^T \boldsymbol\Gamma^2 \textbf{w}_1 - \textbf{w}_2^T \boldsymbol\Gamma^2 \textbf{w}_2 \\
        &= 2\textbf{w}_2^T \boldsymbol\Gamma^2 (\textbf{w}_1 - \textbf{w}_2) + (\textbf{w}_1-\textbf{w}_2)^T \Gamma^2 (\textbf{w}_1-\textbf{w}_2) \\
        \frac{\left\Vert L(\textbf{w}_1) - L(\textbf{w}_2) \right\Vert}{\left\Vert \textbf{w}_1-\textbf{w}_2 \right\Vert} & \leq 2 \left\Vert \textbf{w}_2 \right\Vert \left\Vert \boldsymbol\Gamma^2 \right\Vert + \left\Vert \textbf{w}_1-\textbf{w}_2 \right\Vert \left\Vert \boldsymbol\Gamma^2 \right\Vert 
    \end{aligned}
\]

If $\textbf{w}_1, \textbf{w}_2$ are bounded by $K$, 

\[
    \boxed{
        L = 2K \left\Vert \boldsymbol\Gamma^2 \right\Vert
    }
\]

This additional term may be added to the Lipschitz constants derived above when gradient descent is performed on a loss function including a Tikhonov regularization term. Clearly, for an $L_2$-regularizer, since $\boldsymbol\Gamma = \frac{\lambda}{2}\textbf{I}$, we have $L = \lambda K$.

\subsection{A Note on Sigmoid activation and cross-entropy loss}
Suppose that $y*$ is a latent, continuous random variable. In the case of logistic regression, (leading to binary cross entropy), we assume that 
$y* ~ f(x) + e$ where $e \sim$ logit(mean = 0) pdf. We can see that sigmoid is the CDF of logit distribution (or conversely, sigmoid is a valid CDF, and its derivate is the logit distribution), where $f(x)$ is the mean. Define $y = I(y* >0 )$. Then, 
\begin{align*}
p(y=1) &= p(y*>0) \\
           &= 1- p(f(x) + e <0) \\
           &= 1- \text{cdf}_{\text{logit}}(-f(x)) \\
           &= 1- \sigma(-f(x))
\end{align*}

But $1-\sigma(-x) = \sigma(x)$ due to symmetry. Therefore,

$$ p(y=1) = \sigma(f(x)) $$

The negative log-likelihood under the above generative model results in the binary cross entropy. 
\section{Going Beyond SGD}
\label{sec:otheralg}

The framework presented so far easily extends to algorithms that extend SGD, such as RMSprop, momentum, and Adam. In this section, we show algorithms for some major optimization algorithms popularly used.

RMSprop, gradient descent with momentum, and Adam are based on exponentially weighted averages of the gradients. The trick then is to compute the Lipschitz constant as an exponentially weighted average of the norms of the gradients. This makes sense, since it provides a supremum of the ``velocity" or ``accumulator" terms in momentum and RMSprop respectively.

\subsection{Gradient Descent with Momentum}
SGD with momentum uses an exponentially weighted average of the gradient as a velocity term. The gradient is replaced by the velocity in the weight update rule.

\begin{algorithm}[H]
    \SetAlgoLined
    $K \gets 0$; $V_{\nabla L} \gets 0$\;
    \For{each iteration}{
        Compute $\nabla_W L$ for all layers\;
        $V_{\nabla L} \gets \beta V_{\nabla L} + (1-\beta) \nabla_W L$\;
        \tcp{Compute the exponentially weighted average of LC}
        $K \gets \beta K + (1-\beta) \max \norm{\nabla_W L}$ \label{algo:mom:lc}\;
        \tcp{Weight update}
        $W \gets W - \frac{1}{K}V_{\nabla L}$ \label{algo:mom:update}\;
    }
    \caption{AdaMo}
    \label{algo:mom}
\end{algorithm}

Algorithm \ref{algo:mom} shows the \textit{adaptive} version of gradient descent with momentum. The only changes are on lines \ref{algo:mom:lc} and \ref{algo:mom:update}. The exponentially weighted average of the Lipschitz constant ensures that the learning rate for that iteration is optimal. The weight update is changed to reflect our new learning rate. We use the symbol $W$ to consistently refer to the weights as well as the biases; while ``parameters" may be a more apt term, we use $W$ to stay consistent with literature.

Notice that only line \ref{algo:mom:lc} is our job; deep learning frameworks will typically take care of the rest; we simply need to compute $K$ and use a learning rate scheduler that uses the inverse of this value.

\subsection{RMSprop}
RMSprop uses an exponentially weighted average of the square of the gradients. The square is performed element-wise, and thus preserves dimensions. The update rule in RMSprop replaces the gradient with the ratio of the current gradient and the exponentially moving average. A small value $\epsilon$ is added to the denominator for numerical stability.

Algorithm \ref{algo:rms} shows the modified version of RMSprop. We simply maintain an exponentially weighted average of the Lipschitz constant as before; the learning rate is also replaced by the inverse of the update term, with the exponentially weighted average of the square of the gradient replaced with our computed exponentially weighted average.

\begin{algorithm}[H]
    \SetAlgoLined
    $K \gets 0$; $S_{\nabla L} \gets 0$\;
    \For{each iteration}{
        Compute $\nabla_W L$ on mini-batch\;
        $S_{\nabla L} \gets \beta S_{\nabla L} + (1-\beta) (\nabla_W L)^2$\;
        \tcp{Compute the exponentially weighted average of LC}
        $K \gets \beta K + (1-\beta) \max \norm{(\nabla_W L)^2}$ \label{algo:rms:lc}\;
        \tcp{Weight update}
        $W \gets W - \frac{\sqrt{K} + \epsilon}{\max \norm{\nabla_{W} L}}\cdot \frac{\nabla_{W} L}{\sqrt{S_{\nabla L\textit{}}} + \epsilon}$ \label{algo:rms:update}\;
    }
    \caption{Adaptive RMSprop}
    \label{algo:rms}
\end{algorithm}

\subsection{Adam}
Adam combines the above two algorithms. We thus need to maintain two exponentially weighted average terms. The algorithm, shown in Algorithm \ref{algo:adam}, is quite straightforward.

\begin{algorithm}[H]
    \SetAlgoLined
    $K_1 \gets 0$; $K_2 \gets 0$; $S_{\nabla L} \gets 0$; $V_{\nabla L} = 0$\;
    \For{each iteration}{
        Compute $\nabla_W L$ on mini-batch\;
        $V_{\nabla L} \gets \beta_1 V_{\nabla L} + (1-\beta_1) \nabla_W L$\;
        $S_{\nabla L} \gets \beta_2 S_{\nabla L} + (1-\beta_2) (\nabla_W L)^2$\;
        \tcp{Compute the exponentially weighted averages of LC}
        $K_1 \gets \beta_1 K_1 + (1-\beta_1) \max \norm{\nabla_W L}$ \;
        $K_2 \gets \beta_2 K_2 + (1-\beta_2) \max \norm{(\nabla_W L)^2}$ \;
        \tcp{Weight update}
        $W \gets W - \frac{\sqrt{K_2} + \epsilon}{K_1}\cdot \frac{V_{\nabla L}}{\sqrt{S_{\nabla L\textit{}}} + \epsilon}$ \label{algo:adam:update}\;
    }
    \caption{Auto-Adam}
    \label{algo:adam}
\end{algorithm}

In our experiments, we use the defaults of $\beta_1 = 0.9, \beta_2 = 0.999$.

In practice, it is difficult to get a good estimate of $\max \norm{(\nabla_W L)^2}$. For this reason, we tried two different estimates:
\begin{itemize}
    \item $\norm{(\max \nabla_W L)^2} = \norm{\left( \frac{k-1}{km}K_z + \lambda \norm{w} \right)^2}$ -- This set the learning rate high (around 4 on CIFAR-10 with DenseNet), and the model quickly diverged.
    \item $\left( \max \norm{\nabla_W L} \right)^2 = \frac{(k-1)^2}{k^2 m^2} \max K_z^2 + \lambda^2 (\max \norm{w})^2 + \frac{2\lambda(k-1)}{km}K_z (\max \norm{w})$ -- This turned out to be an overestimation, and while the same model above did not diverge, it oscillated around a local minimum. We fixed this by removing the middle term. This worked quite well empirically.
\end{itemize}

\subsection{A note on bias correction}
Some implementations of the above algorithms perform bias correction as well. This involves computing the exponentially weighted average, and then dividing by $1 - \beta^t$, where $t$ is the epoch number. In this case, the above algorithms may be adjusted by also dividing the Lipschitz constants by the same constant.

\section{Experiments and Results}
\label{sec:experiments}

In this section, we show through extensive experimentation that our approach converges faster, and performs better than with a standard choice of learning rate.

\subsection{Faster convergence}
For checking the rate of convergence, we use classical machine learning models. In each experiment, we randomly initialize weights, and use the same initial weight vector for gradient descent with both the learning rates. In all experiments, we scale each feature to sum to 1 before running gradient descent. This scaled data is used to compute the Lipschitz constants, and consequently, the learning rates. Normalizing the data is particularly important because the Lipschitz constant may get arbitrarily large, thus making the learning rate too small.

The regression experiments use a multiple linear regression model, the binary classification experiments use an ordinary logistic regression model, and the multi-class classification experiments use a softmax regression model with one-hot encoded target labels. For MNIST, however, we found it quicker to train a neural network with only an input and output layer (no hidden layers were used), a stochastic gradient descent optimizer, and softmax activations.

We compare the rate of convergence by setting a threshold, $T_L$ for the value of the loss function. When the value of the cost function goes below this threshold, we stop the gradient descent procedure. A reasonable threshold value is chosen for each dataset separately. We then compare $E_{0.1}$ and $E_{1/L}$, where $E_\alpha$ represents the number of epochs taken for the loss to go below $T_L$. For the Cover Type data, we considered only the first two out of seven classes. This resulted in 495,141 rows. We also considered only ten features to speed up computation time.

For the least-squares cost function, an estimate of $K$ is required. A good estimate of $K$ would be obtained by running gradient descent with some fixed learning rate and then taking the norm of the final weight vectors. However, because this requires actually running the algorithm for which we want to find a parameter first, we need to estimate this value instead. In our experiments, we obtain a close approximation to the value obtained above through the formula below. For the experiments in this subsection, we use this formula to compute $K$.

\[
    \begin{aligned}
        a &= \frac{1}{m}\sum\limits_{j=1}^n \sum\limits_{i=1}^m x^{(i)}_j \\
        b &= \frac{1}{n}\sum\limits_{j=1}^n \max\limits_i x^{(i)}_j \\
        K &= \frac{a+b}{2}
    \end{aligned}
\]
In the above formulas, the notation $x^{(i)}_j$ refers to the $j$th column of the $i$th training example. Note that $a$ is the sum of the means of each column, and $b$ is the mean of the maximum of each column.

\begin{figure}
    \centering
    \includegraphics[scale=0.5]{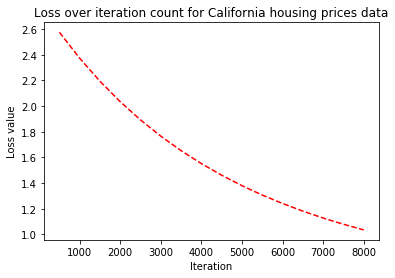}
    \caption{Loss function over iterations for California housing prices dataset}
    \label{fig:leastsq:1}
\end{figure}

\begin{table}
    \caption{Comparison of speed of convergence with $\alpha=0.1$ and $\alpha=\frac{1}{L}$. The speed of convergence under the influence of adaptive learning rate, $\alpha=\frac{1}{L}$ is evident. $L$ is the Lipschitz constant.}
    \centering
    \begin{tabular}{lllll}
        \toprule \\
        Dataset & $T_L$ & $1/L$ & $E_{0.1}$ & $E_{1/L}$ \\
        \midrule \\
        Boston housing prices & 200 & 9.316 & 46,041 & 555 \\
        California housing prices & 2.8051 & 5163.5 & 24,582 & 2 \\
        Energy efficiency \cite{tsanas2012accurate} & 100 & 12.78 & 489,592 & 3,833 \\
        Online news popularity \cite{fernandes2015proactive} & 73,355,000 & 1.462 & 10,985 & 753 \\
        Breast cancer & 0.69 & 4280.23 & 37,008 & 2 \\
        Covertype\footnotemark & 0.69314 & 17.48M & 216,412 & 2 \\
        Iris & 0.2 & 1.902 & 413 & 49 \\
        Digits & 0.2 & 0.634 & 337 & 2 \\
        \bottomrule \\
     \end{tabular}
    \label{tab:exp:1}
\end{table}

\footnotetext{We restricted the data to the first 100K rows only.}

Table \ref{tab:exp:1} shows the results of our experiments on some datasets. Clearly, our choice of $\alpha$ outperforms a random guess in all the datasets. Our proposed method yields a learning rate that adapts to each dataset to converge significantly faster than with a guess. In some datasets, our choice of learning rate gives over a 100x improvement in training time.

While the high learning rates may raise concerns of oscillations rather than convergence, we have checked for this in our experiments. To do this, we continued running gradient descent, monitoring the value of the loss function every 500 iterations. Figure \ref{fig:leastsq:1} shows this plot, demonstrating that the high learning rates indeed lead to convergence.

\subsection{Better performance}
We tested the performance of our approach with both classical machine learning models and deep neural networks. In this section, we discuss the results of the former. To compare, we ran the models with different learning rates for a fixed number of epochs, $N_E$, and compared the accuracy scores $A_{0.1}$ and $A_{1/L}$, where $A_\alpha$ is the accuracy score after $N_E$ epochs with the learning rates 0.1 and $1/L$ respectively.

\begin{table}
    \caption{Comparison of performance with $\alpha=0.1$ and $\alpha=\frac{1}{L}$. $A_{1/L}$ denotes adaptive learning rate proposed in the paper.}
    \centering
    \begin{tabular}{lllll}
        \toprule \\
        Dataset & $N_E$ & $1/L$ & $A_{0.1}$ & $A_{1/L}$ \\
        \midrule \\
        Iris & 200 & 1.936 & 93.33\% & 97.78\% \\
        Digits & 200 & 0.635 & 91.3\% & 94.63\% \\
        MNIST & 200 & 10.24 & 92.7\% & 92.8\% \\
        Cover Type\footnotemark & 1000 & 189.35M & 43.05\% & 57.21\% \\
        Breast cancer & 1000 & 4280.22 & 43.23\% & 90.5\% \\
        \bottomrule \\
     \end{tabular}
    \label{tab:exp:2}
\end{table}

\footnotetext{The inverse Lipschitz constant is different here because the number of rows was not restricted to 100K. Also, the inverse Lipschitz constant here is not a typo. The learning rate was indeed set to 189.35 million.}

\begin{figure}
    \centering
    \includegraphics[scale=0.5]{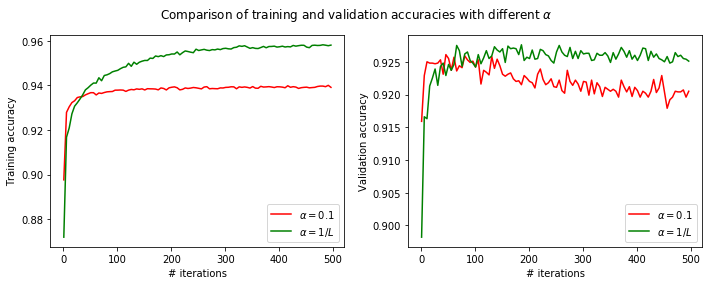}
    \caption{Comparison of training and validation accuracy scores for different $\alpha$ on MNIST}
    \label{fig:classif:1}
\end{figure}

Table \ref{tab:exp:2} shows the results of these experiments. Figure \ref{fig:classif:1} shows a comparative plot of the training and validation accuracy scores for both learning rates on the MNIST dataset. In both the plots, the red line is for $\alpha = 0.1$, while the green line is for $\alpha = \frac{1}{L}$. Although our choice of learning rate starts off worse, it quickly ($< 100$ iterations) outperforms a learning rate of 0.1. Further, the validation accuracy has a \textit{decreasing} tendency for $\alpha = 0.1$, while it is more stable for $\alpha=1/L$.

\subsection{Better performance in deep neural networks}
We compared the performance of our approach with deep neural networks on standard datasets as well. While our results are not state of the art, our focus was to empirically show that optimization algorithms can be run with higher learning rates than typically understood. On CIFAR, we only use flipping and translation augmentation schemes as in \cite{he2016deep}. In all experiments the raw image values were divided by 255 after removing the means across each channel. We also provide baseline experiments performed with a fixed learning rate for a fair comparison, using the same data augmentation scheme.

\begin{longtable}{cccccc}
    \caption{Summary of all experiments: abbreviations used - LR: Learning Rate; WD: weight decay; VA: validation accuracy} \\
        \toprule \\
        Dataset & Architecture & Algorithm & LR Policy & WD & VA. \\
        \midrule \\
        MNIST & Custom & SGD & Adaptive & None & 99.5\% \\
        MNIST & Custom & Momentum & Adaptive & None & \textbf{99.57\%} \\
        MNIST & Custom & Adam & Adaptive & None & 99.43\% \\
        \midrule \\
        CIFAR-10 & ResNet20 & SGD & Baseline & $10^{-3}$ & 60.33\% \\
        CIFAR-10 & ResNet20 & SGD & Fixed & $10^{-3}$ & 87.02\% \\
        CIFAR-10 & ResNet20 & SGD & Adaptive & $10^{-3}$ & 89.37\% \\
        CIFAR-10 & ResNet20 & Momentum & Baseline & $10^{-3}$ & 58.29\% \\
        CIFAR-10 & ResNet20 & Momentum & Adaptive & $10^{-2}$ & 84.71\% \\
        CIFAR-10 & ResNet20 & Momentum & Adaptive & $10^{-3}$ & 89.27\% \\
        CIFAR-10 & ResNet20 & RMSprop & Baseline & $10^{-3}$ & 84.92\% \\
        CIFAR-10 & ResNet20 & RMSprop & Adaptive & $10^{-3}$ & 86.66\% \\
        CIFAR-10 & ResNet20 & Adam & Baseline & $10^{-3}$ & 84.67\%\\
        CIFAR-10 & ResNet20 & Adam & Fixed & $10^{-4}$ & 70.57\% \\
        CIFAR-10 & DenseNet & SGD & Baseline & $10^{-4}$ & 84.84\% \\
        CIFAR-10 & DenseNet & SGD & Adaptive & $10^{-4}$ & 91.34\% \\
        CIFAR-10 & DenseNet & Momentum & Baseline & $10^{-4}$ & 85.50\%\\
        CIFAR-10 & DenseNet & Momentum & Adaptive & $10^{-4}$ & \textbf{92.36\%} \\
        CIFAR-10 & DenseNet & RMSprop & Baseline & $10^{-4}$ & 91.36\% \\
        CIFAR-10 & DenseNet & RMSprop & Adaptive & $10^{-4}$ & 90.14\% \\
        CIFAR-10 & DenseNet & Adam & Baseline & $10^{-4}$ & 91.38\% \\
        CIFAR-10 & DenseNet & Adam & Adaptive & $10^{-4}$ & 88.23\% \\
        \midrule \\
        CIFAR-100 & ResNet56  & SGD & Adaptive & $10^{-3}$ & 54.29\% \\
        CIFAR-100 & ResNet164  & SGD & Baseline & $10^{-4}$ & 26.96\% \\
        CIFAR-100 & ResNet164  & SGD & Adaptive & $10^{-4}$ & \textbf{75.99\%} \\
        CIFAR-100 & ResNet164  & Momentum & Baseline & $10^{-4}$ & 27.51\% \\
        CIFAR-100 & ResNet164  & Momentum & Adaptive & $10^{-4}$ & 75.39\% \\
        CIFAR-100 & ResNet164  & RMSprop & Baseline & $10^{-4}$ & 70.68\% \\
        CIFAR-100 & ResNet164  & RMSprop & Adaptive & $10^{-4}$ & 70.78\% \\
        CIFAR-100 & ResNet164  & Adam & Baseline & $10^{-4}$ & 71.96\% \\
        CIFAR-100 & DenseNet & SGD & Baseline & $10^{-4}$ & 50.53\% \\
        CIFAR-100 & DenseNet & SGD & Adaptive & $10^{-4}$ & 68.18\% \\
        CIFAR-100 & DenseNet & Momentum & Baseline & $10^{-4}$ & 52.28\%\\
        CIFAR-100 & DenseNet & Momentum & Adaptive & $10^{-4}$ & 69.18\% \\
        CIFAR-100 & DenseNet & RMSprop & Baseline & $10^{-4}$ & 65.41\% \\
        CIFAR-100 & DenseNet & RMSprop & Adaptive & $10^{-4}$ & 67.30\% \\
        CIFAR-100 & DenseNet & Adam & Baseline & $10^{-4}$ & 66.05\% \\
        CIFAR-100 & DenseNet & Adam & Adaptive & $10^{-4}$ & 40.14\%\footnotemark \\
        \bottomrule \\
    \label{tab:exp:summary}
\end{longtable}
\footnotetext{This was obtained after 67 epochs. After that, the performance deteriorated, and after 170 epochs, we stopped running the model. We also ran the model on the same architecture, but restricting the number of filters to 12, which yielded 59.08\% validation accuracy.}

A summary of our experiments is given in Table \ref{tab:exp:summary}. DenseNet refers to a DenseNet\cite{huang2017densely} architecture with $L = 40$ and $k = 12$.

\subsubsection{MNIST}
On MNIST, the architecture we used is shown in Table \ref{tab:mnist:1}. All activations except the last layer are ReLU; the last layer uses softmax activations. The model has 730K parameters.

\begin{table}
    \centering
    \caption{CNN used for MNIST}
    \begin{tabular}{ccc}
        \toprule \\
        Layer & Filters & Padding \\
        \midrule \\
        3 x 3 Conv & 32 & Valid  \\
        3 x 3 Conv & 32 & Valid  \\
        2 x 2 MaxPool & -- & -- \\
        Dropout (0.2) & -- & -- \\
        3 x 3 Conv & 64 & Same \\
        3 x 3 Conv & 64 & Same \\
        2 x 2 MaxPool & -- & -- \\
        Dropout (0.25) & -- & -- \\
        3 x 3 Conv & 128 & Same \\
        Dropout (0.25) & -- & -- \\
        Flatten & -- & -- \\
        Dense (128) & -- & -- \\
        BatchNorm & -- & -- \\
        Dropout (0.25) & -- & -- \\
        Dense (10) & -- & -- \\
        \bottomrule \\
    \end{tabular}
    \label{tab:mnist:1}
\end{table}

Our preprocessing involved random shifts (up to 10\%), zoom (to 10\%), and rotations (to $15^\circ$). We used a batch size of 256, and ran the model for 20 epochs. The experiment on MNIST used only an adaptive learning rate, where the Lipschitz constant, and therefore, the learning rate was recomputed every epoch. Note that this works even though the penultimate layer is a Dropout layer. No regularization was used during training. With these settings, we achieved a training accuracy of 98.57\% and validation accuracy 99.5\%.

\begin{figure}
    \centering
    \begin{subfigure}[b]{0.3\textwidth}
        \includegraphics[width=\linewidth]{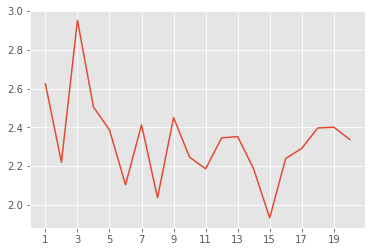}
        \caption{}
        \label{fig:lrhist:a}
    \end{subfigure}
    \begin{subfigure}[b]{0.3\textwidth}
        \includegraphics[width=\linewidth]{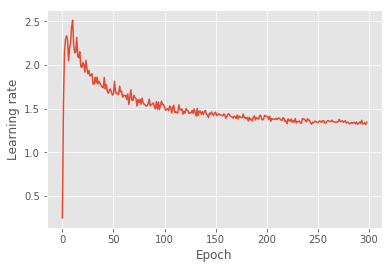}
        \caption{}
        \label{fig:lrhist:b}
    \end{subfigure}
    \begin{subfigure}[b]{0.3\textwidth}
        \includegraphics[width=\linewidth]{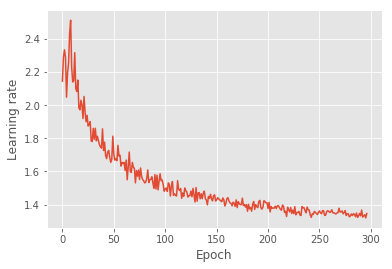}
        \caption{}
        \label{fig:lrhist:c}
    \end{subfigure}
    \begin{subfigure}[b]{0.3\textwidth}
        \includegraphics[width=\linewidth]{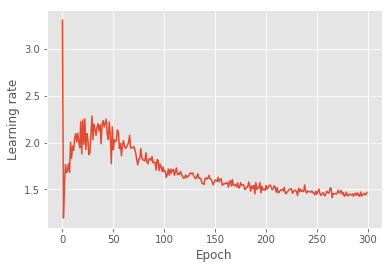}
        \caption{}
        \label{fig:lrhist:d}
    \end{subfigure}
    \begin{subfigure}[b]{0.3\textwidth}
        \includegraphics[width=\linewidth]{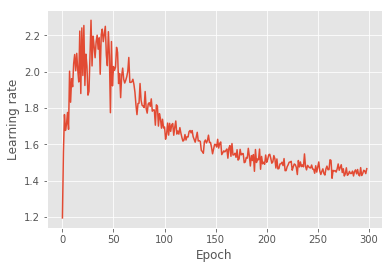}
        \caption{}
        \label{fig:lrhist:e}
    \end{subfigure}
    \begin{subfigure}[b]{0.3\textwidth}
        \includegraphics[width=\linewidth]{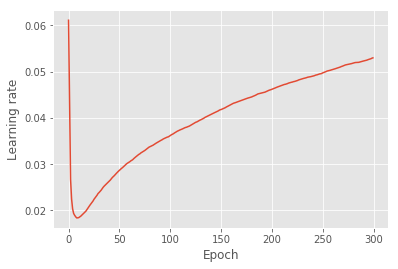}
        \caption{}
        \label{fig:lrhist:f}
    \end{subfigure}
    \begin{subfigure}[b]{0.3\textwidth}
        \includegraphics[width=\linewidth]{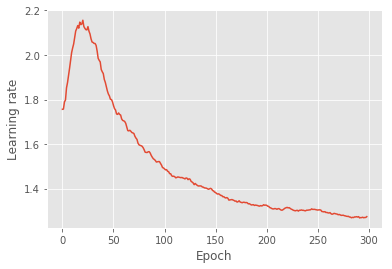}
        \caption{}
        \label{fig:lrhist:g}
    \end{subfigure}
    \begin{subfigure}[b]{0.3\textwidth}
        \includegraphics[width=\linewidth]{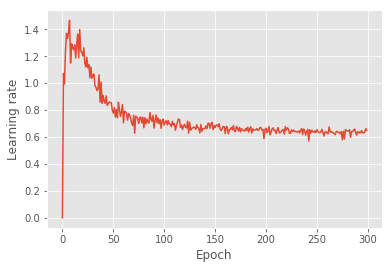}
        \caption{}
        \label{fig:lrhist:h}
    \end{subfigure}
    \begin{subfigure}[b]{0.3\textwidth}
        \includegraphics[width=\linewidth]{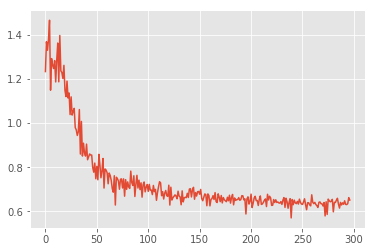}
        \caption{}
        \label{fig:lrhist:i}
    \end{subfigure}
    \caption{Plots of adaptive learning rate over time with various architectures and datasets. Where not specified, an SGD optimizer may be assumed. (a) Custom architecture on MNIST (b) ResNet20 on CIFAR-10 (c) ResNet20 on CIFAR-10 from epoch 2 (d) DenseNet on CIFAR-10 (e) DenseNet on CIFAR-10 from epoch 2 (f) DenseNet on CIFAR-10 with Adam optimizer (g) DenseNet on CIFAR-10 using AdaMo (h) ResNet164 on CIFAR-100 (i) ResNet164 on CIFAR-100 from epoch 3}
    \label{fig:lrhist:all}
\end{figure}


Finally, Figure \ref{fig:lrhist:a} shows the computed learning rate over epochs. Note that unlike the computed adaptive learning rates for CIFAR-10 (Figures \ref{fig:lrhist:b} and \ref{fig:lrhist:c}) and CIFAR-100 (Figures \ref{fig:lrhist:h} and \ref{fig:lrhist:i}), the learning rate for MNIST starts at a much higher value. While the learning rate here seems much more random, it must be noted that this was run for only 20 epochs, and hence any variation is exaggerated in comparison to the other models, run for 300 epochs.

The results of our Adam optimizer is also shown in Table \ref{tab:exp:summary}. The optimizer achieved its peak validation accuracy after only 8 epochs. 

We also used a custom implementation of SGD with momentum (see Appendix \ref{appendix:A} for details), and computed an adaptive learning rate using our AdaMo algorithm. Surprisingly, this outperformed both our adaptive SGD and Auto-Adam algorithms. However, the algorithm consistently chose a large (around 32) learning rate for the first epoch before computing more reasonable learning rates--since this hindered performance, we modified our AdaMo algorithm so that on the first epoch, the algorithm sets $K$ to 0.1 and uses this value as the learning rate. We discuss this issue further in Section \ref{sec:cifar10}.

\subsubsection{CIFAR-10} \label{sec:cifar10}
For the CIFAR-10 experiments, we used a ResNet20 v1\cite{he2016deep}. A residual network is a deep neural network that is made of ``residual blocks". A residual block is a special case of a highway networks \cite{srivastava2015highway} that do not contain any gates in their skip connections. ResNet v2 also uses ``bottleneck" blocks, which consist of a 1x1 layer for reducing dimension, a 3x3 layer, and a 1x1 layer for restoring dimension \cite{he2016identity}. More details can be found in the original ResNet papers \cite{he2016deep,he2016identity}.

We ran two sets of experiments on CIFAR-10 using SGD. First, we empirically computed $K_z$ by running one epoch and finding the activations of the penultimate layer. We ran our model for 300 epochs using the same fixed learning rate. We used a batch size of 128, and a weight decay of $10^{-3}$. Our computed values of $K_z$, $\max \norm{w}$, and learning rate were 206.695, 43.257, and 0.668 respectively. It should be noted that while computing the Lipschitz constant, $m$ in the denominator must be set to the batch size, not the total number of training examples. In our case, we set it to 128. 

\begin{figure}
    \centering
    \includegraphics[scale=0.4]{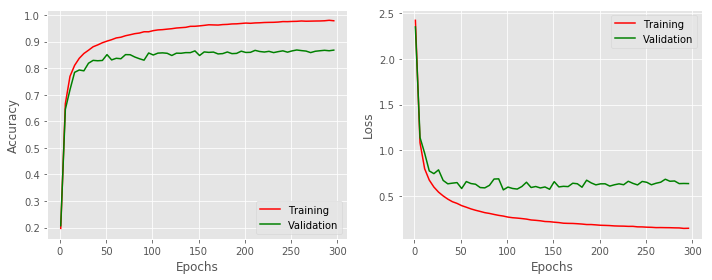}
    \caption{Plot of accuracy score and loss over epochs on CIFAR-10}
    \label{fig:cifar10:1}
\end{figure}

Figure \ref{fig:cifar10:1} shows the plots of accuracy score and loss over time. As noted in \cite{smith2018disciplined}, a horizontal validation loss indicates little overfitting. We achieved a training accuracy of 97.61\% and a  validation accuracy of 87.02\% with these settings.

Second, we used the same hyperparameters as above, but recomputed $K_z$, $\max \norm{w}$, and the learning rate every epoch. We obtained a training accuracy of 99.47\% and validation accuracy of 89.37\%. Clearly, this method is superior to a fixed learning rate policy. 


Figure \ref{fig:lrhist:b} and \ref{fig:lrhist:c} show the learning rate over time. The adaptive scheme automatically chooses a decreasing learning rate, as suggested by literature on the subject. On the first epoch, however, the model chooses a very small learning rate of $8 \times 10^{-3}$, owing to the random initialization. 

Observe that while it does follow the conventional wisdom of choosing a higher learning rate initially to explore the weight space faster and then slowing down as it approaches the global minimum, it ends up choosing a significantly larger learning rate than traditionally used. Clearly, there is no need to decay learning rate by a multiplicative factor. Our model with adaptive learning rate outperforms our model with a fixed learning rate in only 65 epochs. Further, the generalization error is lower with the adaptive learning rate scheme using the same weight decay value. This seems to confirm the notion in \cite{smith2017super} that large learning rates have a regularization effect.


Figures \ref{fig:lrhist:d} and \ref{fig:lrhist:e} show the learning rate over time on CIFAR-10 using a DenseNet architecture and SGD. Evidently, the algorithm automatically adjusts the learning rate as needed.


Interestingly, in all our experiments, ResNets consistently performed poorly when run with our auto-Adam algorithm. Despite using fixed and adaptive learning rates, and several weight decay values, we could not optimize ResNets using auto-Adam. DenseNets and our custom architecture on MNIST, however, had no such issues. Our best results with auto-Adam on ResNet20 and CIFAR-10 were when we continued using the learning rate of the first epoch (around 0.05) for all 300 epochs. 

Figure \ref{fig:lrhist:f} shows a possible explanation. Note that over time, our auto-Adam algorithm causes the learning rate to slowly increase. We postulate that this may be the reason for ResNet's poor performance using our auto-Adam algorithm. However, using SGD, we are able to achieve competitive results for all architectures. We discuss this issue further in Section \ref{sec:practical}.

ResNets did work well with our AdaMo algorithm, though, performing nearly as well as with SGD. As with MNIST, we had to set the initial learning rate to a fixed value with AdaMo. We find that a reasonable choice of this is between 0.1 and 1 (both inclusive). We find that for higher values of weight decay, lower values of $x$ perform better, but we do not perform a more thorough investigation in this paper. In our experiments, we choose $x$ by simply trying 0.1, 0.5, and 1.0, running the model for five epochs, and choosing the one that performs the best. In Table \ref{tab:exp:summary}, for the first experiment using ResNet20 and momentum, we used $x = 0.1$; for the second, we used $x = 1$.


AdaMo also worked well with DenseNets on CIFAR-10. We used $x=0.5$ for this model. This model crossed 90\% validation accuracy before 100 epochs, maintaining a learning rate higher than 1, and was the best among all our models trained on CIFAR-10. This shows the strength of our algorithm. Figure \ref{fig:lrhist:g} shows the learning rate over epochs for this model.

\subsubsection{CIFAR-100}
For the CIFAR-100 experiments, we used a ResNet164 v2 \cite{he2016identity}. Our experiments on CIFAR-100 only used an adaptive learning rate scheme.

We largely used the same parameters as before. Data augmentation involved only flipping and translation. We ran our model for 300 epochs, with a batch size of 128. As in \cite{he2016identity}, we used a weight decay of $10^{-4}$. We achieved a training accuracy of 99.68\% and validation accuracy of 75.99\% with these settings.

For the ResNet164 model trained using AdaMo, we found $x=0.5$ to be the best among the three that we tried. Note that it performs competitively compared to SGD. For DenseNet, we used $x=1$.


Figures \ref{fig:lrhist:h} and \ref{fig:lrhist:i} show the learning rate over epochs. As with CIFAR-10, the first two epochs start off with a very small ($10^{-8}$) learning rate, but the model quickly adjusts to changing weights.

\subsubsection{Baseline Experiments}
For our baseline experiments, we used the same weight decay value as our other experiments; the only difference was that we simply used a fixed value of the default learning rate for that experiment. For SGD and SGD with momentum, this meant a learning rate of 0.01. For Adam and RMSprop, the learning rate was 0.001. In SGD with momentum and RMSprop, $\beta = 0.9$ was used. For Adam, $\beta_1 = 0.9$ and $\beta_2 = 0.999$ were used.

\subsection{Comparison with other methods}

    In this section, we compare our results against other methods proposed in recent years. Because many papers do not test on CIFAR-100, we show comparisons on MNIST and CIFAR-10. We show the number of epochs and the validation accuracy. In cases where the paper reports only the validation loss but not accuracy, we do not make a comparison. We also show our best performing result for comparison in each table.

 For a fair comparison, we compare against the closest architecture used. Below, we compare against several recent adaptive learning rate methods.
    
\textbf{AdaShift \cite{zhou2018adashift}:} The authors show their results on MNIST and CIFAR-10 using a ResNet and a DenseNet. We compare our results accordingly in Tables \ref{tab:adashit-cif10-rn}-\ref{tab:adashit-mnist}.
    
    \begin{table}[h]
        \centering
        \caption{Comparison of ResNets on CIFAR-10 between AdaShift and LipschitzLR.}
        \label{tab:adashit-cif10-rn}
        \begin{tabular}{llll}
        \toprule
            \textbf{Training Details} & \textbf{AdaShift} & \textbf{LipschitzLR} & \textbf{LipschitzLR (custom)} \\
            \midrule
            Architecture & ResNet18 & ResNet 20 & DenseNet-40 \\
            Epochs & 50 & 300 & 300 \\
            Validation accuracy & 91.25\% & 89.37\% & 92.36\% \\
            \bottomrule
        \end{tabular}
    \end{table}
    
    \begin{table}[h]
        \centering
        \caption{Comparison of DenseNets on CIFAR-10 between AdaShift and LipschitzLR. The DenseNet with AdaMo yields our best results on CIFAR-10.}
        \label{tab:adashit-cif10-dn}
        \begin{tabular}{lll}
        \toprule
            \textbf{Training Details} & \textbf{AdaShift} & \textbf{LipschitzLR} \\
            \midrule
            Architecture & 100-layer DenseNet & 40-layer DenseNet \\
            Epochs & 150 & 300 \\
            Validation accuracy & 90\% & 91.34\% (SGD), 92.36\% (AdaMo) \\
            \bottomrule
        \end{tabular}
    \end{table}
    
    \begin{table}[h]
        \centering
        \caption{Comparison of neural networks on MNIST between AdaShift and LipschitzLR.}
        \label{tab:adashit-mnist}
        \begin{tabular}{lll}
        \toprule
            \textbf{Training Details} & \textbf{AdaShift} & \textbf{LipschitzLR} \\
            \midrule
            Architecture & 2 hidden layers & 16 layers as in Table \ref{tab:mnist:1} \\
            Epochs & 20 & 20 \\
            Validation accuracy & 89\% & 99.5\% (SGD), 99.57\% (AdaMo) \\
            \bottomrule
        \end{tabular}
    \end{table}

\textbf{AdaDelta \cite{zeiler2012adadelta}}: The authors show their results on MNIST, and we provide a comparison in Table \ref{tab:adadelta-mnist}.

\begin{table}[h]
        \centering
        \caption{Comparison of networks on MNIST between AdaDelta and LipschitzLR.}
        \label{tab:adadelta-mnist}
        \begin{tabular}{lll}
        \toprule
            \textbf{Training Details} & \textbf{AdaDelta} & \textbf{LipschitzLR} \\
            \midrule
            Architecture & 2-layer network with tanh activations & CNN described in Table \ref{tab:mnist:1} \\
            Epochs & 6 & 20 \\
            Validation accuracy & 98.17\% & 99.5\% (SGD), 99.57\% (AdaMo) \\
            \bottomrule
        \end{tabular}
    \end{table}

\textbf{WN-Grad \cite{wu2018wngrad}}: The authors show their results on MNIST and CIFAR-10, but for MNIST, they do not show validation accuracy scores, only the loss values. However, it is incorrect to compare learners using their loss function values, as different points on the loss surface can have the same value but leading to different accuracy scores. Therefore, we only show a comparison on CIFAR-10 in Table \ref{tab:wngrad-cif10-rn}.

\begin{table}[h]
        \centering
        \caption{Comparison of ResNets on CIFAR-10 between WN-Adam and LipschitzLR.}
        \label{tab:wngrad-cif10-rn}
        \begin{tabular}{llll}
        \toprule
            \textbf{Training Details} & \textbf{WN-Adam} & \textbf{LipschitzLR} & \textbf{LipschitzLR (custom)} \\
            \midrule
            Architecture & ResNet18 & ResNet 20 & DenseNet-40 \\
            Epochs & 100 & 300 & 300 \\
            Validation accuracy & 92\% & 89.37\% & 92.36\% \\
            \bottomrule
        \end{tabular}
    \end{table}
    
\textbf{AMSGrad \cite{reddi2019convergence}}: The authors of AMSGrad only report loss values on MNIST and CIFAR-10, so we find that we cannot directly compare against their results. Fortunately, the authors of AdaShift \cite{zhou2018adashift} also report AMSGrad validation accuracy scores, so we compare our results against those values in Tables \ref{tab:amsgrad-cif10-rn} - \ref{tab:amsgrad-mnist}.

\begin{table}[h]
        \centering
        \caption{Comparison of ResNets on CIFAR-10 between AMSGrad and LipschitzLR.}
        \label{tab:amsgrad-cif10-rn}
        \begin{tabular}{llll}
        \toprule
            \textbf{Training Details} & \textbf{AMSGrad} & \textbf{LipschitzLR} & \textbf{LipschitzLR (custom)} \\
            \midrule
            Architecture & ResNet 18 & ResNet 20 & DenseNet-40 \\
            Epochs & 50 & 300 & 300 \\
            Validation accuracy & 90\% & 89.37\% & 92.36\% \\
            \bottomrule
        \end{tabular}
    \end{table}
    
    \begin{table}[h]
        \centering
        \caption{Comparison of DenseNets on CIFAR-10 between AMSGrad and LipschitzLR.}
        \label{tab:amsgrad-cif10-dn}
        \begin{tabular}{lll}
        \toprule
            \textbf{Training Details} & \textbf{AMSGrad} & \textbf{LipschitzLR} \\
            \midrule
            Architecture & DenseNet-100 & DenseNet-40 \\
            Epochs & 150 & 300 \\
            Validation accuracy & 88.75\% & 91.34\% (SGD), 92.36\% (AdaMo) \\
            \bottomrule
        \end{tabular}
    \end{table}
    
    \begin{table}[h]
        \centering
        \caption{Comparison of networks on MNIST between AMSGrad and LipschitzLR.}
        \label{tab:amsgrad-mnist}
        \begin{tabular}{lll}
        \toprule
            \textbf{Training Details} & \textbf{AMSGrad} & \textbf{LipschitzLR} \\
            \midrule
            Architecture & 2-layer network & CNN described in Table \ref{tab:mnist:1} \\
            Epochs & 20 & 20 \\
            Validation accuracy & 89\% & 99.5\% (SGD), 99.57\% (AdaMo) \\
            \bottomrule
        \end{tabular}
    \end{table}

\section{Practical Considerations} \label{sec:practical}
Although our approach is theoretically sound, there are a few practical issues that need to be considered. In this section, we discuss these issues, and possible remedies.

The first issue is that our approach takes longer per epoch than with choosing a standard learning rate. Our code was based on the Keras deep learning library, which to the best of our knowledge, does not include a mechanism to get outputs of intermediate layers directly. Other libraries like PyTorch, however, do provide this functionality through ``hooks". This eliminates the need to perform a partial forward propagation simply to obtain the penultimate layer activations, and saves computation time. We find that computing $\max \norm{w}$ takes very little time, so it is not important to optimize its computation.

Another issue that causes practical issues is random initialization. Due to the random initialization of weights, it is difficult to compute the correct learning rate for the first epoch, because there is no data from a previous epoch to use. We discussed the effects of this already with respect to our AdaMo algorithm, and we believe this is the reason for the poor performance of auto-Adam in all our experiments. Fortunately, if this is the case, it can be spotted within the first two epochs--if large values of the intermediate computations: $\max \norm{w}$, $K_z$, etc. are observed, then it may be required to set the initial LR to a suitable value. We discussed this for the AdaMo algorithm. In practice, we find that for RMSprop, this rarely occurs; but when it does, the large intermediate values are shown in the very first epoch. We find that a small value like $10^{-3}$ works well as the initial LR. In our experiments, we only had to do this for ResNet on CIFAR-100.

\section{Discussion and Conclusion}
In this paper, we derived a theoretical framework for computing an adaptive learning rate; on deriving the formulas for various common loss functions, it was revealed that this is also ``adaptive" with respect to the data. We explored the effectiveness of this approach on several public datasets, with commonly used architectures and various types of layers.

Clearly, our approach works ``out of the box" with various regularization methods including $L_2$, dropout, and batch normalization; thus, it does not interfere with regularization methods, and automatically chooses an optimal learning rate in stochastic gradient descent. On the contrary, we contend that our computed larger learning rates do indeed, as pointed out in \cite{smith2017super}, have a regularizing effect; for this reason, our experiments used small values of weight decay. Indeed, increasing the weight decay significantly hampered performance. This shows that ``large" learning rates may not be harmful as once thought; rather, a large value may be used if carefully computed, along with a guarded value of $L_2$ weight decay. We also demonstrated the efficacy of our approach with other optimization algorithms, namely, SGD with momentum, RMSprop, and Adam. In summary, we have shown, beyond reasonable doubt, that sufficiently deep architectures may not be a necessity to accomplish state-of-the-art performance.

Our auto-Adam algorithm performs surprisingly poorly. We postulate that like AdaMo, our auto-Adam algorithm will perform better when initialized more thoughtfully. To test this hypothesis, we re-ran the experiment with ResNet20 on CIFAR-10, using the same weight decay. We fixed the value of $K_1$ to 1, and found the best value of $K_2$ in the same manner as for AdaMo, but this time, checking $10^{-3}$, $10^{-4}$, $10^{-5}$, and $10^{-6}$. We found that the lower this value, the better our results, and we chose $K_2 = 10^{-6}$. While at this stage we can only conjecture that this combination of $K_1$ and $K_2$ will work in all cases, we leave a more thorough investigation as future work. Using this configuration, we achieved 83.64\% validation accuracy. 

A second avenue of future work involves obtaining a tighter bound on the Lipschitz constant and thus computing a more accurate learning rate. Another possible direction is to investigate possible relationships between the weight decay and the initial learning rate in the AdaMo algorithm.

\section*{Conflicts of Interest}
\textbf{Funding:} This work was supported by the Science and Engineering Research Board (SERB)-Department of Science and Technology
(DST), Government of India (project reference number SERB-EMR/ 2016/005687). The funding source was not involved in the study design, writing of the report, or in the decision to submit this article for publication.

The authors declare that they have no other conflicts of interest.

\appendix
\section{Implementation Details} \label{appendix:A}
All our code was written using the Keras deep learning library. The architecture we used for MNIST was taken from a Kaggle Python notebook by Aditya Soni\footnote{https://www.kaggle.com/adityaecdrid/mnist-with-keras-for-beginners-99457}. For ResNets, we used the code from the Examples section of the Keras documentation\footnote{https://keras.io/examples/cifar10\_resnet/}. The DenseNet implementation we used was from a GitHub repository by Somshubra Majumdar\footnote{https://github.com/titu1994/DenseNet}. Finally, our implementation of SGD with momentum is a modified version of the Adam implementation in Keras\footnote{https://github.com/keras-team/keras/blob/master/keras/optimizers.py\#L436}.

\bibliographystyle{splncs04}
\bibliography{cite}

\end{document}